\documentclass{article}

\pdfoutput=1

\usepackage[preprint,nonatbib]{neurips_2022}

\usepackage[utf8]{inputenc} 
\usepackage{graphicx}
\usepackage{float}
\usepackage{wrapfig}
\usepackage{amsmath}
\usepackage{bbm}
\usepackage{algorithm}
\usepackage{algorithmic}
\usepackage[T1]{fontenc}    
\usepackage{hyperref}       
\usepackage{url}            
\usepackage{booktabs}       
\usepackage{amsfonts}       
\usepackage{nicefrac}       
\usepackage{microtype}      
\usepackage{xcolor}         
\usepackage{biblatex} 
\usepackage{caption}

\title{Memory-Based Label-Text Tuning for Few-Shot Class-Incremental Learning}

\author{%
  Jinze Li \\
  School of Integrated Circuits \\
  University of Chinese Academy of Sciences \\
  Beijing, China \\
  \texttt{lijinze20@mails.ucas.ac.cn} \\
  \And
  Yan Bai \\
  School of Computer Science \\
  Peking University \\
  Beijing, China \\
  \texttt{yanbai@pku.edu.cn} \\
  \And
  Yihang Lou \\
  Intelligent Vision Dept \\
  Huawei Technologies \\
  Beijing, China \\
  \texttt{louyihang1@huawei.com} \\
  \And
  Xiongkun Linghu \\
  Department of Electronic Engineering \\
  Tsinghua University \\
  Beijing, China \\
  \texttt{lhxk20@mails.tsinghua.edu.cn} \\
  \AND
  Jianzhong He \\
  Intelligent Vision Dept \\
  Huawei Technologies \\
  Beijing, China \\
  \texttt{jianzhong.he@huawei.com} \\
  \And
  Shaoyun Xu \\
  Institute of Microelectronics \\
  Chinese Academy of Sciences \\
  Beijing, China \\
  \texttt{xushaoyun@ime.ac.cn} \\
  \And
  Tao Bai \\
  Intelligent Vision Dept \\
  Huawei Technologies \\
  Beijing, China \\
  \texttt{baitao13@huawei.com} \\
}
   
\DeclareUnicodeCharacter{0301}{\'{e}}
\begin{document}

\maketitle

\begin{abstract}
Few-shot class-incremental learning(FSCIL) focuses on designing learning algorithms that can continually learn a sequence of new tasks from a few samples without forgetting old ones. The difficulties are that training on a sequence of limited data from new tasks leads to severe overfitting issues and causes the well-known catastrophic forgetting problem. Existing researches mainly utilize the image information, such as storing the image knowledge of previous tasks or limiting classifiers updating. However, they ignore analyzing the informative and less noisy text information of class labels. In this work, we propose leveraging the label-text information by adopting the memory prompt. The memory prompt can learn new data sequentially, and meanwhile store the previous knowledge. Furthermore, to optimize the memory prompt without undermining the stored knowledge, we propose a stimulation-based training strategy. It optimizes the memory prompt depending on the image embedding stimulation, which is the distribution of the image embedding elements. Experiments show that our proposed method outperforms all prior state-of-the-art approaches, significantly mitigating the catastrophic forgetting and overfitting problems. 
\end{abstract}

\section{Introduction}
Deep neural networks have achieved remarkable success in various computer vision tasks\cite{resnet, imagenet, vit}. With sufficient data and massive computational resources at the training stage, the model usually provides a satisfactory result on the pre-defined image categories by a fine-tuning method. However, it is impossible to get sufficient data on all the classes at the training stage in real-world %
scenarios\cite{few_shot_survey}. It is more likely to get a limited amount of class-specific data incrementally as time goes by, which is termed few-shot class-incremental learning(FSCIL)\cite{fscil}. Suppose a pre-trained model is directly fine-tuned on the new class training set. Then two severe problems arise (1) over-fitting\cite{fscil}, which means the fine-tuned model overfits on the limited training data and thus fails to maintain satisfying generalization,
and (2) catastrophic forgetting\cite{1989_catastrophic, catastrophic_forgetting1, ewc}, which means the model suffers a dramatic performance drop on previous classes.

To solve the problems mentioned above, many approaches are proposed. 
Rehearsal and replay based approaches\cite{icarl, end_to_end, rebalancing, gradiant_based_sample_selection, c_fscil} save the exemplar sets of previous tasks and use them to learn the current task jointly. Parameters penalization based approaches\cite{synaptic_intelligence, ewc, memory_aware_synapses} impose constraints on some parameters of the trained model during the training stage to prevent the model from changing to a large extent. Knowledge distillation approaches\cite{less_forgetting_learning, lwf, adaptive_feature_consolidation} aim to minimize the gap between the previous and current tasks. Model expansion approaches\cite{adaptive_aggregation_network, der, progressive_neural_network} present to use a dynamic model instead of a static one to tackle the growing number of classes. 
However, existing methods have shortcomings: (1) Approaches such as rehearsal and model expansion suffer the continuously growing memory consumption with the increase of new classes. (2) The other approaches, such as parameters penalization and knowledge distillation, are more dependent on the similarity between the sequence of tasks\cite{lwf}.  

\begin{wrapfigure}{r}{0.5\textwidth}
    \centering
    \includegraphics[width=0.50\textwidth]{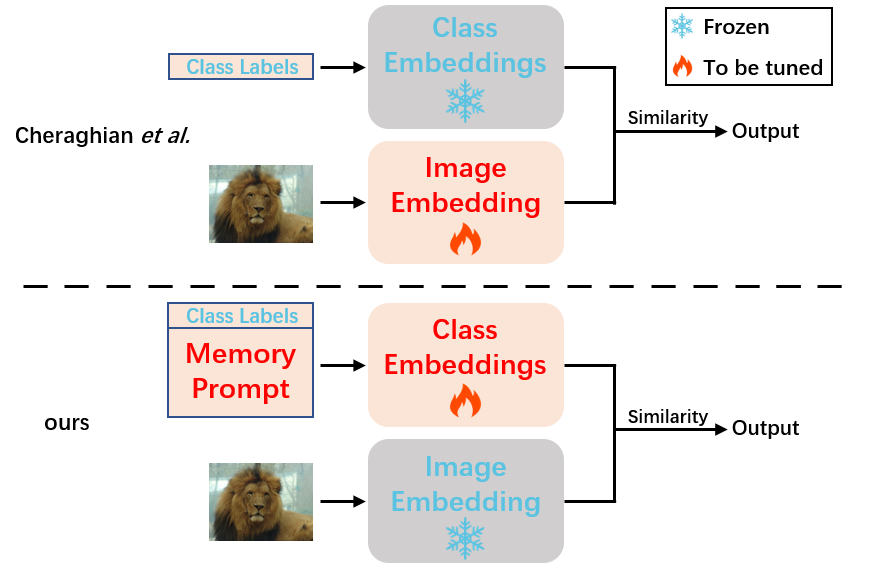}
    \caption{Difference between Cheraghian \emph{et al.}\cite{semantic_aware_knowledge_distillation} and our proposed method. To perform classification by the similarity between image and class label embeddings, Cheraghian \emph{et al.} optimizes the image encoder while we introduce the memory prompt and optimizes the class label embeddings by training the prompt.} 
    \label{compare}
\vspace{-10pt}
\end{wrapfigure}

More importantly, most approaches ignore utilizing the text information of class labels, which contains much information and less noise compared with image information. 
A recent work\cite{semantic_aware_knowledge_distillation} tries to use the fixed text information of class labels as prototypes, and classifies images via the similarity between the image embeddings and the text prototype, as shown in Fig \ref{compare}. 
However, text information can be utilized more effectively. It has the characteristic of lightweight and less noisy compared with image information. Hence it is more reasonable to optimize the text information of image labels. That is to say, we can classify images and store learned knowledge mainly based on the images' label-text embeddings.


In this paper, we propose a memory-based few-shot class-incremental learner(M-FSCIL), which utilizes the lightweight and less noisy text information of class labels together with image information to perform classification. M-FSCIL is architecturally composed of a memory prompt, an image encoder, and a memory interpreter. 
We collect all the learned category label texts and encode them to the word embeddings. Then, we concatenate the memory prompt with each word embedding, and input these concatenations to the memory interpreter. The obtained memory-class concatenation embeddings contain the text information of each category, which we called class label embeddings. Meanwhile, we also input images to the image encoder and get the image embeddings.  
Finally, the model calculates the similarity between the image embeddings and each class label embedding and then classifies the images based on the similarity. 
The novelty of M-FSCIL is the utilization of the label-text information by optimizing the memory prompt. The prompt associate the image features with the label-text information, which concentrates on semantic information with fewer parameters to optimize, and thus mitigate the catastrophic forgetting problem.

Furthermore, to optimize the memory prompt in M-FSCIL, we propose a training strategy based on the image embedding stimulation. The novel strategy trains the memory prompt depending on the image embedding elements. 
It considers the relatively large embedding elements as stimulation, and updates the memory prompt selectively according to the stimulation from the current and previous tasks. The proposed stimulation-based training strategy enables the memory prompt to learn new tasks and maintain stored knowledge simultaneously.
After training with the proposed strategy, these optimized memory prompt elements work as learned knowledge. And these elements are unlikely to update substantially in the subsequent sessions. In a word, we train the memory prompt of M-FSCIL mainly depending on the stimulation from each image embedding element. 

The main contributions of this paper are as follow:
\begin{itemize}
    \item [(1)]
    We propose an M-FSCIL that can more effectively utilize the less-noisy text information of class labels to mitigate the catastrophic forgetting and overfitting problems. 
    \item [(2)]
    Memory prompt is introduced to learn new tasks and remember old ones, which alleviates the growing memory consumption problem.
    \item [(3)]
    We propose a novel stimulation-based training strategy, which optimizes the memory prompt to possess good generalizability on the new tasks without forgetting the old knowledge.
    \item [(4)]
    Experiments on MiniImageNet, CUB200, and CIFAR100 demonstrate the effectiveness of our method and achieve new state-of-the-art results.
\end{itemize}

\section{Related work}
\label{Related work}
\paragraph{Class-incremental learning(CIL).}
The model performs class-incremental learning means continually learning a sequence of tasks. The newly trained model can generalize well currently and not forget the previously learned old tasks. Machine learning community shows great interest on this field, and a great number of researches have emerged recently\cite{ewc, synaptic_intelligence, memory_aware_synapses, lwf, less_forgetting_learning, icarl, gradiant_based_sample_selection, der, adaptive_aggregation_network, progressive_neural_network, deep_generate_replay}. Rebuffi \emph{et al.}\cite{icarl} proposed a rehearsal and replay approach that summarizes the previous tasks by storing some exemplar sets and using them to learn a new task. Kirkpatrick \emph{et al.}\cite{ewc} introduced a parameter penalization approach that prevents models from changing too much from the previously learned ones. 


\paragraph{Few-shot learning.}
The strong performance of deep learning models heavily relies on the abundant labelled data at the training stage, which is unrealistic in a real-world scenario\cite{few_shot_survey}. Thus, the ability of a model to generalize after seeing a limited amount of training data has attracted considerable attention\cite{siamese_network, relation_network, prototypical_network, matching_network, imprinted_weights, dynamic_lwf, tadam}. There are two mainstream approaches that we can learn from, which are the metric-based methods\cite{siamese_network, relation_network, prototypical_network, matching_network} and the optimization-based methods\cite{imprinted_weights, dynamic_lwf, maml}. Metric-based methods aim to use a pre-trained backbone to encode training data of each class and calculate the distance(e.g., L2 distance) to classify the testing data. Optimization-based methods focus on the learning paradigms that enable the model to adapt to different classes given limited training data.

\paragraph{Few-shot class-incremental learning(FSCIL).} Following CIL, FSCIL focus on the challenging problem of learning the streaming new tasks with few samples provided. The restriction of training data exacerbates the over-fitting and catastrophic forgetting problem. Very recently, several works have been proposed in this field\cite{flat_minima, fscil, attention_attractor_network, mgsvf, vector_quantization, semantic_aware_knowledge_distillation, few_shot_lll, cec}. Shi \emph{et al.}\cite{flat_minima} introduced to restrict the parameters updating at the base session and fine-tune the model parameters within the flat local minima of the base training objective function. Tao \emph{et al.}\cite{fscil} proposed a neural gas network to preserve the topology of the feature manifold formed by different classes. Hewsche \emph{et al.}\cite{c_fscil} proposed C-FSCIL that uses hyperdimensional embeddings with the fixed dimensions in the vector space to express many more classes. Zhang \emph{et al.}\cite{cec} proposed a continually evolved classifier that employs a graph model to propagate context information between classifiers for adaptation.

\paragraph{Prompt learning.}
Prompt learning is introduced to overcome the gap between the pre-training model and downstream tasks. Utilizing the prompt can be seen as injecting additional parameters into the pre-trained model and conducting a knowledge probing\cite{knowledge_bases, commonsense_reasoning, commonsense_knowledge_mining}. Some researches show that adopting trainable vectors as prompt are less sensitive and more effective in generalizing compared with word prompt\cite{compare_prmopts1}. Word prompt is called hard prompt, while vector prompt is called soft prompt. 
Similar to Zhou \emph{et al.}\cite{coop}, we show that soft prompt tends to probe the knowledge in the wording embedding space to the representation of images, while we adopt a more sophisticated training strategy, which is capable of the few-shot class-incremental learning task settings.

\section{Proposed method}
In this section, we first introduce the notations and preliminaries of task settings in few-shot class-incremental learning(FSCIL). Then, we describe our proposed memory-based few-shot class-incremental learner(M-FSCIL) and stimulation-based training strategy in detail. Our proposed M-FSCIL exploits the informative category labels to classify images. Meanwhile, proposed stimulation-based training strategy optimizes the M-FSCIL to generalize well on the current session without catastrophic forgetting problems.

\subsection{Notations and Preliminaries}
FSCIL aims to design a learning algorithm that can sequentially learn from only a few new training data without forgetting knowledge learned previously. Suppose a sequence of training tasks $\mathcal{D}=\{\mathcal{D}_1, \mathcal{D}_2, ...,\mathcal{D}_T\}$, and testing tasks $\widetilde{\mathcal{D}}=\{\widetilde{\mathcal{D}_1}, \widetilde{\mathcal{D}_2}, ...,\widetilde{\mathcal{D}_T}\}$. To be more specific, in training task $\mathcal{D}_t = \{(x^{(t)}_n, y^{(t)}_n)\}_{n=1}^{|\mathcal{D}_t|}$, the input data $x^{(t)}_n$ is an image and $y^{(t)}_n$ is the corresponding ground-truth label. The labels $y^{(t)}_n \in \mathcal{C}_t$, where $\mathcal{C}_t$ is the set of classes in the task $\mathcal{D}_t$. Usually, the training sets are denoted as sessions, and the first training session is denoted as the base session, which contains a larger number of training data. And in contrast, the subsequent sessions, which are often present as an N-way K-shot task, only have a few N classes with a limited number of K  samples per class. When the model steps into the next session, the data of previous sessions are no longer accessible and we can only use the current session's data to train the model. The classes in each session have no overlap, $i.e.$,  $\mathcal{C}_i \bigcap \mathcal{C}_j = \phi$ , $\forall i, j \in \{1,...,T\}$ where $i \ne j$. For a testing set $\widetilde{\mathcal{D}_t}$, its classes set $\widetilde{\mathcal{C}_t}$ contains samples from all previous sessions as well as the current one, $i.e.$, $\widetilde{\mathcal{C}_t} = \bigcup_{i=1}^t \mathcal{C}_{i}$. 
For example, the benchmark dataset CIFAR100 contains 60 classes in the base session and 5 classes(5-way) in each subsequent class-incremental session. There are 500 training samples for each class in the base session and 5 images(5-shot) for every class in the incremental sessions. According to the incremental sessions training setting, we call it a 5-way 5-shot task. During the testing stage, we use all of the classes learned before(60 + 5 $\times$ (n-1), where n is the session number) to test the network.

\subsection{M-FSCIL}
\paragraph{Overview.}
\begin{figure}
    \centering
    \includegraphics[width=0.95\textwidth]{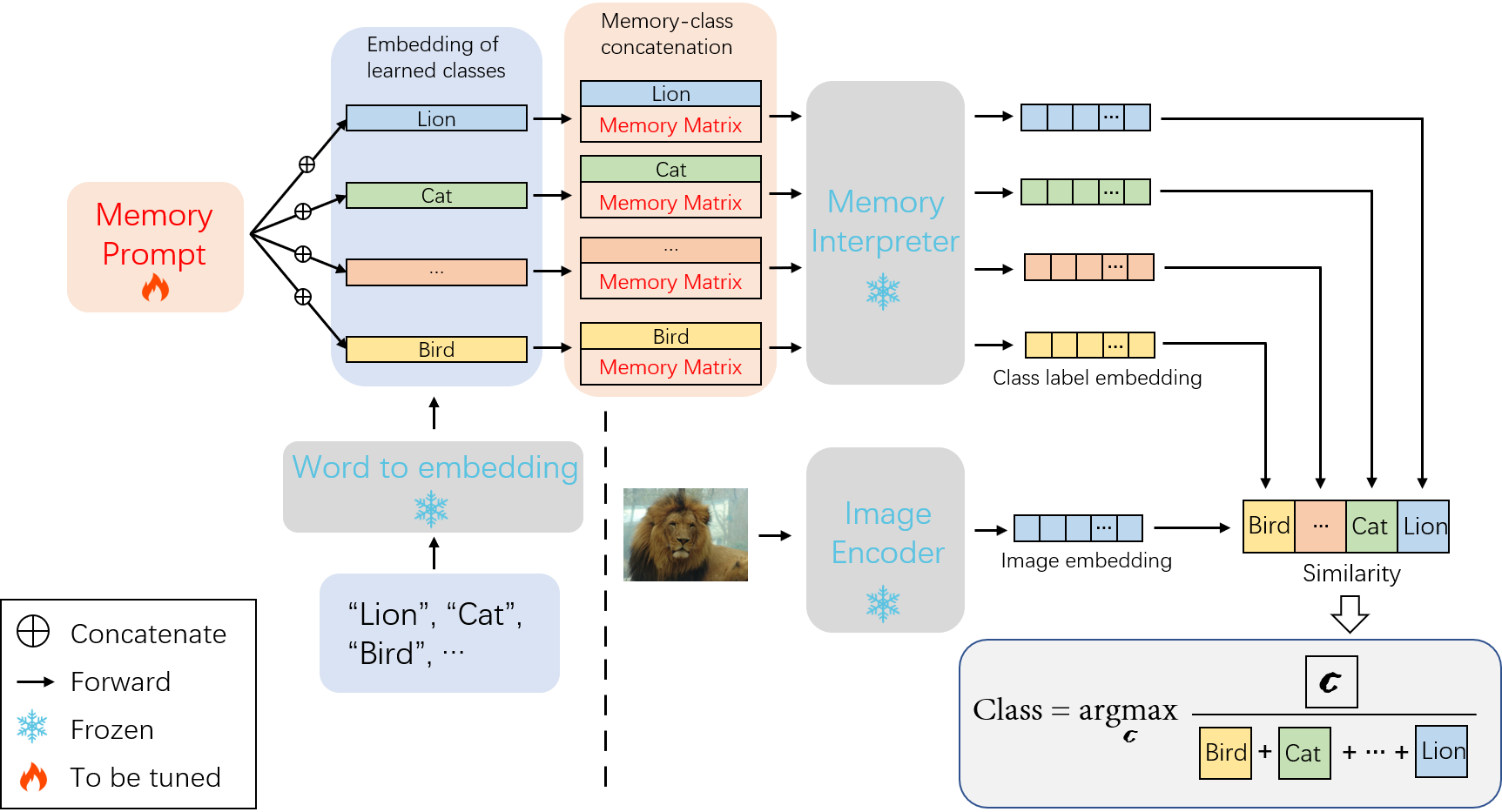}
    \caption{Illustration of our proposed M-FSCIL. To leverage the text information of class labels, we encode the learned labels to embeddings and concatenate them with the trainable memory prompt. And then input the concatenation to the memory interpreter to get the class-specific embeddings. Finally, we calculate the similarity between the image embeddings and each class-specific embedding to classify the input picture.} 
    \label{forward}
\vspace{-8pt}
\end{figure}
Our proposed M-FSCIL can fully utilize the text information of category labels and use the memory prompt to store the learned knowledge. 
The architecture of the network is illustrated in Figure \ref{forward}. It contains three major parts. \textbf{(1) The memory-class concatenation part} includes concatenations of the memory prompt and all the text embeddings of learned classes, \textit{i.e.}, classes in the base and incremental learned sessions.
The class label embeddings are generated by the \emph{word to embedding} part with the input of all learned labels. And the memory prompt is the learnable parameter in M-FSCIL, storing the knowledge learned in the past sessions.
\textbf{(2) The encoder part} contains an image encoder and a memory interpreter. During the training and testing stage, we input the images to the image encoder and then get the image embeddings. Meanwhile, we input the memory-class concatenations to the memory interpreter, which returns the class label embeddings. 
We adopt ViT as image encoder and Bert as memory interpreter, both of which are the transformer-encoder structure, and we use the pre-trained parameters provided by\cite{clip}, which makes full use of the dual-modal information. \textbf{(3) The Similarity Calculation part} calculates the similarity between the image embeddings and all memory-class concatenation embeddings and adopts the most significant similarity one as the predicted class of this image.

\paragraph{Initialization.}
The memory prompt is randomly initialized, and all of the parameters in the prompt are trainable. 
The frozen parameters of \emph{word to embedding}, \emph{memory interpreter} and \emph{image encoder} are pre-trained by the WIT dataset, which is consist of various image-text pairs to train multi-modal networks\cite{clip}.

\paragraph{Embeddings Generation.} 
As shown in Fig. \ref{forward}, we input the training images to the image encoder and get the image embeddings. For the text branch, we input the memory-class concatenation to the memory interpreter and receive the class label embeddings. 
M-FSCIL separates the model's capability into feature extraction and characteristics learning, 
where the \emph{image encoder} part is mainly responsible for extracting features from different images
and the \emph{memory prompt and memory interpreter} part accounts for learning the characteristics of each new category as well as storing the previous knowledge. 

To describe the generation process in detail, we use $\mathcal{G}$ to represent the image encoder, which is responsible for calculating the latent features $\mathcal{I}_i^{(t)}$ of each image $x^{(t)}_i \in \mathcal{D}_t$, where $\mathcal{D}_t$ represents the training set in session $t$. The well-trained image encoder can map images of different categories to separated regions in the high-dimensional feature space: 
\begin{equation}
    \mathcal{I}_i^{(t)} = \mathcal{G}(x^{(t)}_i), \text{where}  \ x^{(t)}_i \in \mathcal{D}_t.
\end{equation}

In the same way, we input the memory-class concatenation $(\Theta;  \widetilde{\mathcal{C}_t})$ to the memory interpreter $\mathcal{F}$ and get the class label embeddings $\mathcal{M}^{(t)}$, where $\Theta$ is the parameters of the memory prompt to be optimized, and $\widetilde{\mathcal{C}_t}$ is the testing class set of the current session. 
By constraining similarity between image embeddings and class label embeddings, the memory prompt can learn the semantic meaning of the images. 
The class label embeddings calculating process can be described as follow:
\begin{equation}
    \mathcal{M}^{(t)} = \mathcal{F}(\Theta; \widetilde{\mathcal{C}_t}).
\end{equation}
The memory interpreter generates the embeddings of all classes in $\widetilde{\mathcal{C}_t}$ simultaneously. In this way, the model can calculate the similarity between the image embeddings with all learned class label embeddings to perform classification.

\paragraph{Inference.} %
After training the M-FSCIL, the corresponding elements of embeddings generated by the image encoder and memory interpreter have a meaningful correlation, so we calculated the cosine similarity of them to indicate the classification results in session $t$:

\begin{align}
    class_{i}^t = \mathop{argmax}\limits_{c}\frac{s_{i,c}}{\sum_{c \in \widetilde{\mathcal{C}_t}} s_{i,c}}, \text{where} \
    s_{i,c} = \frac{\mathcal{I}_i^{(t)} \cdot \mathcal{M}^{(t)}}{||\mathcal{I}_i^{(t)}||_2 \times ||\mathcal{M}^{(t)}||_2}, 
\end{align}

where $i$ is the index of samples,  $s_{i,c}$ indicates the similarity between sample $i$ and class $c$, and $|| \cdot ||_2$ is L2 norm of the vector. 
All of the image embeddings and class label embeddings are 512 dimensions. During the inference stage, the image encoder produces the embeddings of every sequentially input image while the memory interpreter simultaneously generates all the class-specific embeddings learned in the previous and current sessions. 

\subsection{Stimulation-based training strategy}
We introduce a stimulation-based training strategy to optimize the proposed memory prompt. With this strategy, our proposed M-FSCIL can generalize well in the current session and remember the previous knowledge.  
\begin{figure}
    \centering
    \includegraphics[width=0.80\textwidth]{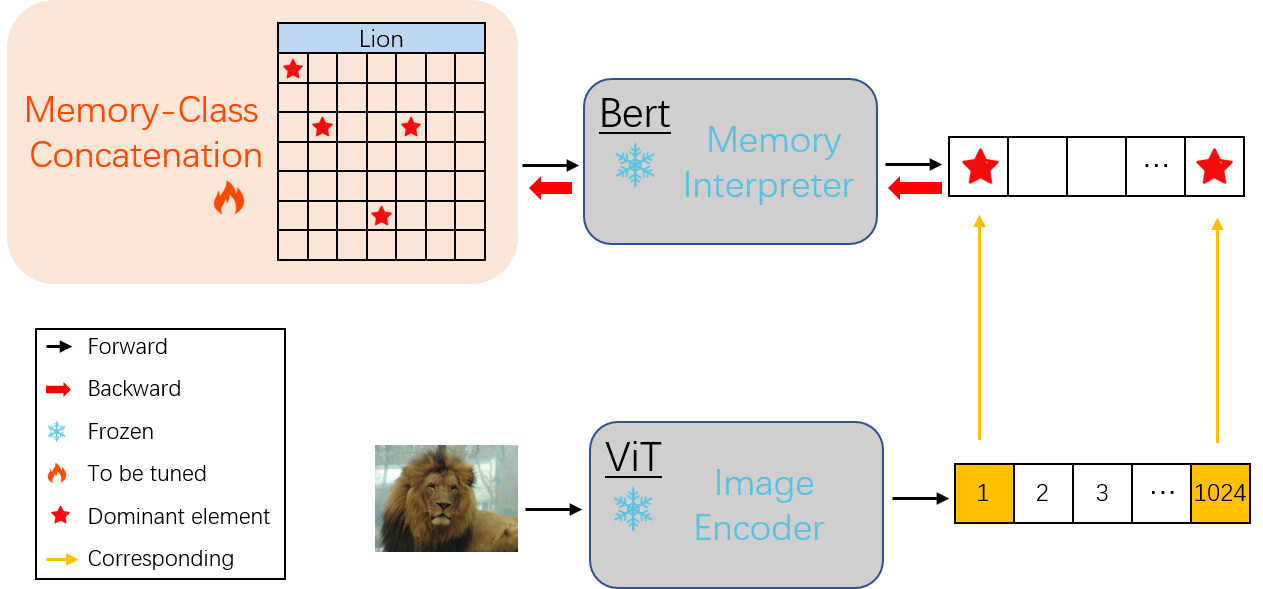}
    \caption{Illustration of Stimulation-based training strategy. The stimulation from image embeddings affects the corresponding elements in the class label embeddings and thus trains the parameters in the memory prompt. By this means, the memory prompt parameters are associated with the image category information.} 
    \label{backward}
\end{figure}

\paragraph{Why do image embeddings contain category information?}
We consider the memory prompt as the storage of knowledge and update it mainly depending on the stimulation from the elements of image embeddings $\mathcal{I}_i^{(t)}$, which is the output of the image encoder.
The image encoder integrates information across the entire image by the self-attention\cite{attention} operations.
As the image encoding proceeds, the model pays more attention to certain image regions which contain more semantic information for classification\cite{vit}. 
Since they have more semantic information about the images, it is more self-relevant within certain regions, so the self-attention value of these regions is larger. 
Hence certain elements in the image embeddings tend to be significant, and the distribution of these relatively large elements has a semantic meaning in classification. Our ultimate goal is to associate the large elements' distribution with the memory prompt parameters.

\paragraph{How does memory prompt learn and store the knowledge?}
As shown in Fig \ref{backward}, some elements of the image embeddings are relatively large, so the same elements of the corresponding class label embeddings tend to be dominant when doing similarity calculations. In other words, these dominant elements determine whether this class label embedding is similar to this image embedding. Hence they act as the key to identifying the objects' classes\cite{vit}. 
We hope that the specific parameters in the memory prompt, which generate the dominating elements, are relatively fixed and unlikely to change significantly in the following learning sessions. 
In this way, after learning a session, the memory has a reasonable understanding of the current session's classes, so the memory prompt parameters we fixed previously are rarely updated in the following sessions to prevent previous knowledge from forgetting.

\paragraph{How does our proposed strategy optimize the M-FSCIL?}
To formally state the idea from the view of the loss function, we assume that the labelled data $d \in \mathcal{D}$, and $d \sim P(d)$, where $P(d)$ is the distribution of training data. $\mathcal{L}$ is the loss function. Our ultimate target is to minimize the expected loss function on the entire training data:
\begin{equation}
    R(\theta) = \int_{\mathcal{R}^{d}} \mathcal{L}(d,\Theta) \ dP(d), 
\end{equation}
where $\Theta$ is the memory prompt, which is the only parameter we need to optimize.
Since it is hard to minimize the expected loss, we reduce its estimation, the ideal empirical loss. We choose to optimize the loss as follows:
\begin{equation}
\label{loss1}
    \mathcal{L}(\Theta)=\frac{1}{|\mathcal{D}|} \sum_{t=1}^{T}(\frac{1}{|\mathcal{D}_{t}|} \sum_{i=1}^{|\mathcal{D}_{t}|} \mathcal{L}_{sess}(d^{(t)}_i,\Theta) + \Gamma \frac{\alpha}{|\Theta|} \sum_{\theta \in \Theta} ||\theta - \theta^*||_2^2, 
\end{equation}

\begin{equation}
    \mathcal{L}_{sess} = - \sum \limits_{c \in \mathcal{C}_{t}} \mathbbm{1}(y^{t}=c^{t})log(\frac{e^{s_{i,c}}}{\sum_{c \in \widetilde{\mathcal{C}_t}}e^{s_{i,c}}}), 
\end{equation}

where $T$ is the entire sessions of the training process. $\theta$ is the elements of $\Theta$, and $\theta^*$ is the elements finishing the training of the base session.
Then, a formal definition of \emph{Signal Stimulation Rate} $\Gamma$ is given as follows.

\paragraph{Definition 1}(Signal Stimulation Rate). Given an image embedding $\mathcal{I}$ and a class label embedding $\mathcal{M}$ generated by a memory interpreter $\mathcal{F}$, the Signal Stimulation Rate $\Gamma$ is:
\begin{equation}
    \Gamma = \mathcal{I} \  \odot \ \bigtriangledown_\Theta \mathcal{M}, 
\end{equation}
if the following conditions are satisfied: (1) The dimension of image embeddings $\mathcal{I}$ and memory interpreter $\mathcal{F}$ are the same. (2) The model is derivable for the input parameters.

Where $\mathcal{I}$ and $\mathcal{M}$ represent the image embeddings and class label embeddings, respectively. $\bigtriangledown_\Theta$ means the derivative of $\Theta$, and $\odot$ indicates the element-wise product.

The signal stimulation rate $\Gamma$ indicates the extent of visual signal stimulation on the parameters $\Theta$. From the empirical loss function $\mathcal{L}(\Theta)$, we can conclude that if the virtual stimulation of a particular channel is relatively large, the corresponding memory prompt elements are more reluctant to change, like a monument in the movement.

\begin{algorithm}
\label{algori}
\caption{memory-based incremental learning. $\mathcal{C}_t$ is classes in each session; $K$ is samples provided in each category; $\mathcal{G}$ is a pre-trained image encoder; $\mathcal{F}$ is a pre-trained memory interpreter; $\gamma$ is the learning rate. We present the following learning steps for our proposed method for a C-way, K-shot incremental learning task.}
\begin{algorithmic} 
\REQUIRE training data $\mathcal{D}_t$ in each session $t$
\ENSURE the memory prompt $\Theta$ which stores the knowledge of all sessions
\STATE \textbf{Initialization:}
\STATE \quad $\Theta \leftarrow$ randomly initialization
\STATE \quad $\Gamma_{0}^1=0 \leftarrow$  initialize $\Gamma$ in base session before training
\WHILE{$t \in [1,T]$}
\STATE $\{(x^{(t)}_n, y^{(t)}_n)\}$ $\leftarrow$ prepare the input data
\STATE $\mathcal{I}^{(t)}=\mathcal{G}(x^{(t)})$ $\leftarrow$ calculate image embeddings by the pre-trained encoder
\STATE $\mathcal{M}^{(t)}=\mathcal{F}(\Theta;\mathcal{C}_t)$ $\leftarrow$ calculate class label embeddings by the pre-trained interpreter
\STATE $\mathcal{L}(\Theta)$ $\leftarrow$ compute the empirical classification loss by Eq. \ref{loss1}
\STATE $\Theta = \Theta- \gamma \bigtriangledown \mathcal{L}(\Theta)$ $\leftarrow$ update the memory prompt
\IF {\emph{finish using all training data in $\mathcal{D}_t$}}
\WHILE{$i \in [1,\mathcal{C}_t]$}
\STATE $\{(x^{(t)}_n, y^{(t)}_n)\} \in \mathcal{D}_t \leftarrow$ review one sample for every class in the current session 
\STATE $\Gamma_{i}^t = \Gamma_{i-1}^t + \mathcal{I}_i \  \otimes \ \bigtriangledown_\Theta \mathcal{M}_i \leftarrow$ calculate virtual signal stimulation rate $\Gamma$ on trained classes to aware which elements in $\Theta$ contain knowledge thus fix it in next sessions
\ENDWHILE
\ENDIF
\ENDWHILE
\end{algorithmic}
\end{algorithm}

The whole training process is described in Algorithm 1. The only parameters we need to optimize in the training sessions are the memory prompt $\Theta$ in our proposed approach, and we update it in a stimulation-based strategy. The simplified expression shows as follows:
\begin{equation}
    \Theta_{t+1} = \Theta_{t} - \gamma \bigtriangledown \mathcal{L}(\Theta), 
\end{equation}
where $\gamma$ is the learning rate of training processes, and $\mathcal{L}(\Theta)$ is the empirical loss which can be calculated by Eq \ref{loss1}

\section{Experiments}
We evaluate our M-FSCIL on three popular few-shot class-incremental learning benchmark datasets, including miniImageNet\cite{miniimagenet}, CIFAR100\cite{cifar100} and Caltech-USCD Birds-200-2011(CUB200)\cite{cub2000}. 

\subsection{Experimental Setup}

During training sessions, We follow the setting in\cite{fscil} to train our models. 
\textbf{(1)} For CIFAR100 and miniImageNet, we choose 60 classes in the base session and the remaining 40 classes in the incremental sessions. In each incremental session, we select 5 classes for training, and each class has 5 samples, which is a 5-way 5-shot training task setting. After each training procedure, all of the classes learned before, i.e., classes in the base and the incremental learned sessions, are used to test the accuracy performance. We have 9 training sessions(1 base and 8 new) totally. \textbf{(2)} For CUB200, we select 100 classes in the base session and the remaining 100 classes in the incremental sessions. In each incremental session, we adopt the 10-way 5-shot task setting, i.e., 10 classes for training and each class has 5 samples. And in the same way, we use the test images of all learned categories to evaluate the accuracy performance after each session.

We adopt the ViT-B/32\cite{vit} as the backbone of the image encoder and adopt the Bert-B\cite{bert} as our memory interpreter.
We use the pre-trained parameters provided by\cite{clip}. 
We utilize the SGD optimizer in all experiments, where the learning rate and batch size are set to 0.02 and 256, respectively. And the epoch numbers for the base session and new sessions are 50 and 100, respectively. We adopt a matrix of $16 \times 512$ size as the memory prompt, where 512 is the fixed size equals to word embeddings. We further analyze the impact of memory prompt scales in the ablation study.

\subsection{Experimental Results Analysis} 
We compare our proposed method with 6 other methods: iCaRL\cite{icarl}, TOPIC\cite{fscil}, IDLVQ-C\cite{vector_quantization}, F2M\cite{flat_minima}, CEC\cite{cec}, C-FSCIL\cite{c_fscil} and FACT\cite{fact}. 
Overall, our proposed M-FSCIL outperforms all state-of-the-art methods on these three benchmark datasets.Because our model's backbone is pre-trained by various image-text pairs, it performs better on the object classification dataset, such as miniImageNet. Images in miniImageNet possess more semantic meanings. In contrast, fine-grained datasets such as CUB200 contain more detailed information and less semantic meaning, which are less suitable for our network. To validate our viewpoint, we perform another experiment that adds more meaningful text to the class labels. Specifically, we add the string \emph{a type of bird} to each class label of the CUB200 dataset.  
For example, we replace the class label \emph{Black footed Albatross} with \emph{Black footed Albatross, a type of bird}.
By doing so, the accuracy performance increases by about 2 \% for each session, and the model converges more rapidly.

\begin{table}
 \vspace{-15pt}
  \caption{Comparative results on the miniImageNet dataset for a 5-way 5-shot class incremental learning. In each session, we test the accuracy on all learned classes.} 
  \label{miniimagenet_table}
  \centering \footnotesize
  \begin{tabular}{lccccccccc}
    \toprule
    & \multicolumn{9}{c}{\textbf{Sessions}}  \\
    \cmidrule(l){2-10}
    {\textbf{Models}} &1  & 2 & 3 & 4 &5 &6 &7 &8 &9  \\
    \midrule
    iCaRL\cite{icarl}       &61.31&46.32&42.94&37.63&30.49&24.00&20.89&18.80&17.21   \\
    TOPIC\cite{fscil}       &61.31&50.09&45.17&41.16&37.48&35.52&32.19&29.46&24.42 \\
    IDLVQ-C\cite{vector_quantization}     &64.77&59.87&55.93&52.62&49.88&47.55&44.83&43.14&41.84             \\
    F2M\cite{flat_minima}         &67.28&63.80&60.38&57.06&54.08&51.39&48.82&46.58&44.65        \\
    CEC\cite{cec}         &72.00&66.83&62.97&59.43&56.70&53.73&51.19&49.24&47.63        \\
    C-FSCIL Mode3\cite{c_fscil}  &76.40&71.14&66.46&63.29&60.42&57.46&54.78&53.11&51.41\\
    FACT\cite{fact}  &72.56&69.63&66.38&62.77&60.6&57.33&54.34&52.16&50.49\\
    M-FSCIL(ours)  & \textbf{93.45} & \textbf{91.82} &\textbf{87.09} &\textbf{88.07} &\textbf{86.75} &\textbf{87.15} & \textbf{85.68} & \textbf{84.80} & \textbf{85.37} \\
    
    \bottomrule
  \end{tabular}
  \vspace{-10pt}
\end{table}

\begin{table}
 \vspace{-8pt}
  \caption{Comparative results on the CIFAR100 dataset for a 5-way 5-shot class incremental learning. In each session, we test the accuracy on all learned classes.} 
  \label{cifar100_table}
  \centering \footnotesize
  \begin{tabular}{lccccccccc}
    \toprule
    & \multicolumn{9}{c}{\textbf{Sessions}}  \\
    \cmidrule(l){2-10}
    {\textbf{Models}} &1  & 2 & 3 & 4 &5 &6 &7 &8 &9  \\
    \midrule
    iCaRL\cite{icarl}       &66.52&57.26&54.27&50.62&47.33&44.99&43.14&41.16&39.49   \\
    TOPIC\cite{fscil}       &64.10&55.88&47.07&45.16&40.11&36.38&33.96&31.55&29.37 \\
    F2M\cite{flat_minima}         &64.71&62.05&59.01&55.58&52.55&49.96&48.08&46.67&44.67        \\
    CEC\cite{cec}         &73.07&68.88&65.26&61.19&58.09&55.57&53.22&51.34&49.14        \\
    C-FSCIL Mode3\cite{c_fscil}  &77.47&72.40&67.47&63.25&59.84&56.95&54.42&52.47&50.47\\
    FACT\cite{fact} &74.60&72.09&67.56&63.52&61.38&58.36&56.28&54.24&52.10\\
    M-FSCIL(ours)  &\textbf{85.55}&\textbf{80.94}&\textbf{77.27}&\textbf{73.51}&\textbf{69.16}&\textbf{66.44}&\textbf{62.01}&\textbf{59.04}&\textbf{55.06} \\
    \bottomrule
  \end{tabular}
   \vspace{-15pt}
\end{table}

\begin{table}
 \vspace{-10pt}
  \caption{Comparative results on the CUB200 dataset for a 10-way 5-shot class incremental learning. In each session, we test the accuracy on all learned classes.} 
  \label{cub200_table}
  \centering \footnotesize
  \resizebox{\textwidth}{14mm}{
  \begin{tabular}{lccccccccccc}
    \toprule
    & \multicolumn{11}{c}{\textbf{Sessions}}  \\
    \cmidrule(l){2-12}
    {\textbf{Models}} &1  & 2 & 3 & 4 &5 &6 &7 &8 &9 &10 &11 \\
    \midrule
    iCaRL\cite{icarl}       &68.68&52.65&48.61&44.16&36.62&29.52&27.83&26.26&24.01&23.89&21.16   \\
    TOPIC\cite{fscil}       &68.68&62.49&54.81&49.99&45.25&41.40&38.35&35.36&32.22&28.31&26.28 \\
    F2M\cite{flat_minima}         &\textbf{81.07}&78.16&75.57&72.89&70.86&68.17&67.01&\textbf{65.26}&63.36&61.76&60.26        \\
    CEC\cite{cec}         &75.85&71.94&68.50&63.50&62.43&58.27&57.73&55.81&54.83&53.52&52.28     \\
    M-FSCIL(ours) &81.04&\textbf{79.73}&\textbf{76.62}&\textbf{73.30}&\textbf{71.22}&\textbf{68.90}&\textbf{66.87}&65.02&\textbf{63.90}&\textbf{62.49}&\textbf{60.40}\\
    \bottomrule
  \end{tabular}}
 \vspace{-15pt}
\end{table}

\subsection{Ablation Study}


\paragraph{Ablation study on the model scale.} In this part, we evaluate whether the improvements obtained are from the expanded model scale. We adopt the pre-trained ViT-L/16 and ViT-L/32\cite{vit} as the backbone for comparison, both of which are larger than the image encoder in M-FSCIL. For a fair comparison, we conduct this ablation experiment on the CIFAR100 dataset, because the ViT-L/16 and ViT-L/32 are pre-trained on the ImageNet\cite{imagenet}, which contains the miniImageNet dataset. As shown in Table \ref{scale_table}, the results indicate that the larger models have a better performance in the base session, which classifies the images with sufficient training samples. However, it suffers the catastrophic forgetting problems in the subsequent sessions. In contrast, our smaller model outperforms them in the final session. This experiment proves that the state-of-the-art performance mostly depends on the utilization of label-text information with the novel training strategy.

\begin{table}[h]
\vspace{-10pt}
\caption{Ablation study of the \textbf{model scale} for 5-way 5-shot incremental learning on CIFAR100.}
\label{scale_table}
\centering \footnotesize
\begin{tabular}{ccc}
    \toprule
    & \multicolumn{2}{c}{\textbf{Sessions}}  \\
    \cmidrule(l){2-3}
    {\textbf{Models}} &Base session  & Final session  \\
    \midrule
    ViT-L/16\cite{vit}       & \textbf{92.80}&15.65  \\
    ViT-L/32\cite{vit}       & 91.80&15.08  \\
    M-FSCIL(ours)  & 85.55& \textbf{51.36}  \\
    \bottomrule
  \end{tabular}
\vspace{-8pt}
\end{table}

\paragraph{Ablation study on the scales of memory prompt.}
The memory prompt can be set in different scales, so in this part, we analyze the impact of memory prompt length on the accuracy performance. We adopt the length of 2, 4, 8, and 16 in this ablation experiment. Table \ref{length_table} shows the results on the miniImageNet dataset. It can be seen that with the increase of the memory prompt length, the model can generalize well in the base session and remember more knowledge in the subsequent sessions.

\begin{table}[h]
\vspace{-10pt}
\caption{Ablation study of the \textbf{memory prompt length L} for 5-way 5-shot incremental learning on miniImageNet.}
\label{length_table}
\centering \footnotesize
\begin{tabular}{lcccc}
    \toprule
     & \multicolumn{4}{c}{\textbf{Memory prompt length}}                  \\
    \cmidrule(l){2-5}
    {\textbf{Session}} & L=2  & L=4 & L=8 & L=16\\
    \midrule
    Session 1 (60 \  \ classes)& 92.63 & 92.90 & 93.15 & \textbf{93.45}    \\
    Session 9 (100 classes)& 64.22  & 79.60 & 81.73 & \textbf{85.37}    \\
    Average \  \   (all \ \  sessions) & 84.03 & 85.65 & 85.68 & \textbf{87.80} \\
    \bottomrule
  \end{tabular}
\vspace{-8pt}
\end{table}

\paragraph{Ablation study on the proposed training strategy.}In this part, we analyze the impact of the proposed stimulation-based training strategy. We train the M-FSCIL with/without the proposed strategy and present results in Table \ref{strategy_table}. 
It indicates that with the memory prompt becoming more extensive, the proposed training strategy has a more significant effect. When the length of memory prompt L=2, experimental results show that the strategy becomes a hindrance because insufficient memory prompt fails to store the previous knowledge and meanwhile learn new tasks.

\begin{table}[h]
\vspace{-10pt}
\caption{Ablation study of the \textbf{training strategy} for 5-way 5-shot incremental learning on miniImageNet.}
\label{strategy_table}
\centering \footnotesize
\begin{tabular}{lcccc}\\  
\toprule
& \multicolumn{4}{c}{\textbf{Testing accuracy in final session}}  \\
\cmidrule(l){2-5}
{\textbf{Training strategy}}&L=2&L=4&L=8&L=16 \\
\midrule
w/ strategy&64.22&\textbf{79.60}&\textbf{81.73}&\textbf{85.37}    \\
w/o strategy&\textbf{72.41}&79.51&77.49&80.82     \\
$\Delta$&-8.19&+0.09&+4.24&+4.55  \\
\bottomrule
\end{tabular}
\vspace{-15pt}
\end{table}

\section{Conclusion}
In this paper, we propose a memory-based text-tuning few-shot class-incremental learner(M-FSCIL), which fully leverages the text information of category labels to mitigate the catastrophic forgetting and overfitting problems. We introduce the memory prompt to use the class-text information of samples. Furthermore, we propose a stimulation based training strategy, which optimizes the memory prompt to generalize well on the current task without forgetting previous knowledge. The experimental results validate the contributions, and our M-FSCIL outperforms state-of-the-art approaches on all the benchmark datasets.


\printbibliography
\end{document}